\documentclass[review,3p,times,11pt,authoryear]{elsarticle}

\usepackage{enumitem}
\usepackage{lineno}
\usepackage[colorlinks,linkcolor=red,anchorcolor=blue,citecolor=green]{hyperref}
\usepackage{multirow}
\usepackage{siunitx}
\usepackage{rotating}
\usepackage{epstopdf}
\usepackage{graphicx}
\usepackage{subfigure}

\usepackage{epstopdf}
\usepackage{times}
\usepackage{longtable}

\usepackage{makeidx}
\usepackage{multirow}
\usepackage{multicol}
\usepackage[dvipsnames,svgnames,table]{xcolor}
\usepackage{graphicx}
\usepackage{epstopdf}
\usepackage{ulem}
\usepackage{url}

\usepackage{array}
\usepackage{amsmath,amssymb,amsfonts}
\usepackage{algorithmic}
\usepackage{algorithm}
\usepackage{graphicx}
\usepackage{textcomp}

\usepackage{epsfig}
\usepackage{makecell}
\usepackage{longtable,booktabs}
\usepackage{longtable}

\usepackage{lscape}

\usepackage[export]{adjustbox}

\usepackage{natbib}
\usepackage[skip=0.333\baselineskip]{caption}

\usepackage{amsmath}

\begin{document}
\begin{frontmatter}
	
	
		
	\title{Simple and Robust Forecasting of Spatiotemporally Correlated Small Earth Data with A Tabular Foundation Model}


\author[label1]{Yuting Yang}
\author[label1]{Gang Mei\corref{cor1}}
\ead{gang.mei@cugb.edu.cn}
\cortext[cor1]{Corresponding author}
\author[label1]{Zhengjing Ma}
\author[label1]{Nengxiong Xu}
\author[label1,label2]{Jianbing Peng}

\address[label1]{School of Engineering and Technology, China University of Geosciences (Beijing), 100083, Beijing, China}
\address[label2]{School of Geological Engineering and Geomatics, Chang'an University, Xi'an, 710064, China}

\begin{abstract}
	
Small Earth data are geoscience observations with limited short-term monitoring variability, providing sparse but meaningful measurements, typically exhibiting spatiotemporal correlations. Spatiotemporal forecasting on such data is crucial for understanding geoscientific processes despite their small scale. However, conventional deep learning models for spatiotemporal forecasting requires task-specific training for different scenarios. Foundation models do not need task-specific training, but they often exhibit forecasting bias toward the global mean of the pretraining distribution. Here we propose a simple and robust approach for spatiotemporally correlated small Earth data forecasting. The essential idea is to characterize and quantify spatiotemporal patterns of small Earth data and then utilize tabular foundation models for accurate forecasting across different scenarios. Comparative results across three typical scenarios demonstrate that our forecasting approach achieves superior accuracy compared to the graph deep learning model (T-GCN) and tabular foundation model (TabPFN) in the majority of instances, exhibiting stronger robustness.

\end{abstract}

\begin{keyword}
    Small Earth data \sep  Spatiotemporal correlations \sep Tabular foundation model \sep Forecasting \sep Deep learning
\end{keyword}
\end{frontmatter}




\section{Introduction}
Small Earth data refers to geoscience time-series observations in which short-term monitoring provides limited informative variation, resulting in only sparse but meaningful measurements being available. These data predominantly possess spatiotemporal correlations. Despite their small scale, forecasting on such data is of critical importance for understanding geoscientific processes \citep{RN1,RN2}. Under limited data conditions, the core challenge of this problem involves the joint modeling of multiple spatially distributed observation stations, effectively capturing temporal dynamics and spatial correlations to enhance the accuracy of future state forecasting.

Small Earth data typically exhibit complex spatiotemporal correlations \citep{RN3}, where observations are correlated not only with their past values but also with neighboring locations. Such spatially correlated time-series data are common in various Earth science disciplines—including meteorology, hydrology, and geology, and exhibit significant spatial dependence and temporal evolution characteristics. Such data include precipitation, temperature, soil moisture, groundwater levels, vegetation indices, and geological disaster events such as landslides, often obtained through remote sensing imagery, ground-based monitoring networks, or model simulations \citep{RN4,RN5}. 

Traditional time-series forecasting has primarily used statistical models like Autoregressive Integrated Moving Average (ARIMA) \citep{RN6}, Kalman Filter \citep{RN7}, and Exponential Smoothing \citep{RN8} for univariate linear time-series forecasting. These methods face significant limitations in geospatial applications due to assumptions of data stationarity and inability to model complex spatial dependencies, resulting in poor performance on non-linear, non-stationary, and spatially correlated datasets. Deep learning has enabled significant advances in high-precision forecasting applications, with LSTMs and GRUs successfully applied to geoscience time-series forecasting \citep{RN2,RN9}, offering superior nonlinear modeling and long-term dependency capture compared to traditional methods. However, these conventional approaches predominantly employ univariate time-series forecasting and fail to account for the spatial correlations that are ubiquitous among multivariate time-series in geoscientific datasets.

Forecasting approaches for spatiotemporal data have been developed to better characterize spatial dependencies and temporal dynamics in small Earth data. A prominent example is the use of Temporal Graph Convolutional Network (T-GCN) \citep{RN11}, which effectively captures interactions between spatial topological structures and temporal sequences, demonstrating superior performance in traffic flow forecasting, air quality monitoring, and power load forecasting. Forecasting approaches for spatiotemporal data based on T-GCN have been actively explored in geoscientific applications, including deep learning approaches based on Graph Convolutional Networks (GCN) that combine spatial graph convolution with temporal sequence modeling for landslide deformation displacement forecasting, significantly improving forecasting accuracy for multivariate time series data \citep{RN12}. GCN-based deep learning models (TGCN-GRU, TGCN-LSTM, Attention-TGCN) have been developed to comprehensively consider spatiotemporal characteristics of displacement data, enhancing landslide displacement forecasting performance \citep{RN13}. Moreover, geometric deep learning-based spatiotemporal Graph Convolutional Network models have been introduced for short-term precipitation forecasting, which automatically learn interactions between grid cells and extract spatial relationships through GCN layers and temporal information through 1D convolutional layers, achieving favorable forecasting results on radar reflectivity data \citep{RN14}.

However, T-GCN based forecasting approaches for spatiotemporally correlated small Earth data exhibit significant limitations. These approaches were developed for large-scale, dense datasets and their performance is compromised in geoscientific applications with limited data scale and sparse sampling. Moreover, different application scenarios necessitate tailored model design and extensive retraining, thereby limiting robustness and imposing task-specific efforts for each new scenario.

In recent years, the foundation models have revolutionized deep learning across diverse domains, demonstrating remarkable capabilities in natural language processing, computer vision, and scientific applications through large-scale pre-training and transfer learning paradigms \citep{RN15,RN16,RN17}. These foundation models, characterized by their ability to leverage vast amounts of data during pre-training and subsequently adapt to downstream tasks with minimal fine-tuning, have shown particular promise in addressing data scarcity and generalization challenges. Building upon this foundation model paradigm, a novel meta-learning approach—the Tabular Prior-data Fitted Network (TabPFN) \citep{RN18}—has been proposed to address the aforementioned challenges. TabPFN is characterized as a Bayesian approach that enables zero-shot learning, facilitating rapid forecasting for time-series tasks in structured tabular data without requiring hyperparameter tuning or model retraining. The model is implemented as a large neural network that has been trained on extensive datasets, thereby demonstrating exceptional adaptability and versatility across diverse applications. The core advantage of TabPFN is attributed to its prior data fitting capability, which is built upon the Transformer architecture and enables rapid adaptation to new data through context-learning mechanisms without necessitating task-specific retraining. In addition, the approach exhibits robustness to small sample sizes and offers simplicity of operation. The latest TabPFN-TS \citep{RN19} integrates TabPFN-v2 with lightweight feature engineering to achieve both point forecasting and probabilistic forecasting for time-series problems.

However, several limitations have been identified in the TabPFN based approaches. While these foundation models do not require task-specific training, they often exhibit forecasting bias toward the global mean of the pretraining distribution. Since TabPFN are designed to simulate an extremely large prior distribution for forecasting tasks, their capacity to learn domain-specific geoscientific knowledge is constrained, and the absence of adaptation to small Earth data characteristics prevents these approaches from achieving optimal forecasting accuracy in specialized geospatial forecasting scenarios.

In summary, there are several challenges in forecasting of spatiotemporally correlated small Earth data. First, T-GCN based approaches require extensive historical observational data and involve complex model construction, necessitating tailored model design and extensive retraining for different application scenarios, thereby limiting robustness. Second, while existing foundation models demonstrate simplicity and enable zero-shot forecasting without task-specific training, they often exhibit forecasting bias toward the global mean of the pretraining distribution and lack adequate capacity for learning domain-specific geoscientific knowledge. These models demonstrate insufficient characterization and quantification for spatiotemporal correlations of small Earth data, consequently limiting their ability to achieve superior predictive performance in specialized geospatial forecasting scenarios.

Here we propose a simple and robust approach for forecasting of spatiotemporally correlated small Earth data. The key idea behind our approach is to characterize and quantify spatiotemporal patterns by extracting inherent spatial and temporal features across different scenarios and employ tabular foundation model as the core forecasting architecture, thereby establishing an effective forecasting approach for spatiotemporally correlated small Earth data forecasting. We demonstrate the effectiveness of our approach through comprehensive experiments using diverse geoscience datasets, including precipitation data from five meteorological stations around Xining City along the Qinghai-Tibet Railway \citep{RN20}, displacement data from the Baige landslide \citep{RN21,RN22},, and surface moisture data from the Qingmu Guan watershed in the southwestern mountainous region \citep{RN23}. This approach reduces development costs while improving accuracy, achieving efficient and accurate spatiotemporally correlated small Earth data forecasting and providing novel methodological support for geoscience data analysis. Our findings offer valuable insights into the application of tabular foundation models in scientific domains and promote further development in geoscience spatiotemporal forecasting.
 
\section{ Materials and methods }
\subsection{Data collection and process}

\begin{figure}[!ht]
	\centering
	\includegraphics[width=16cm]{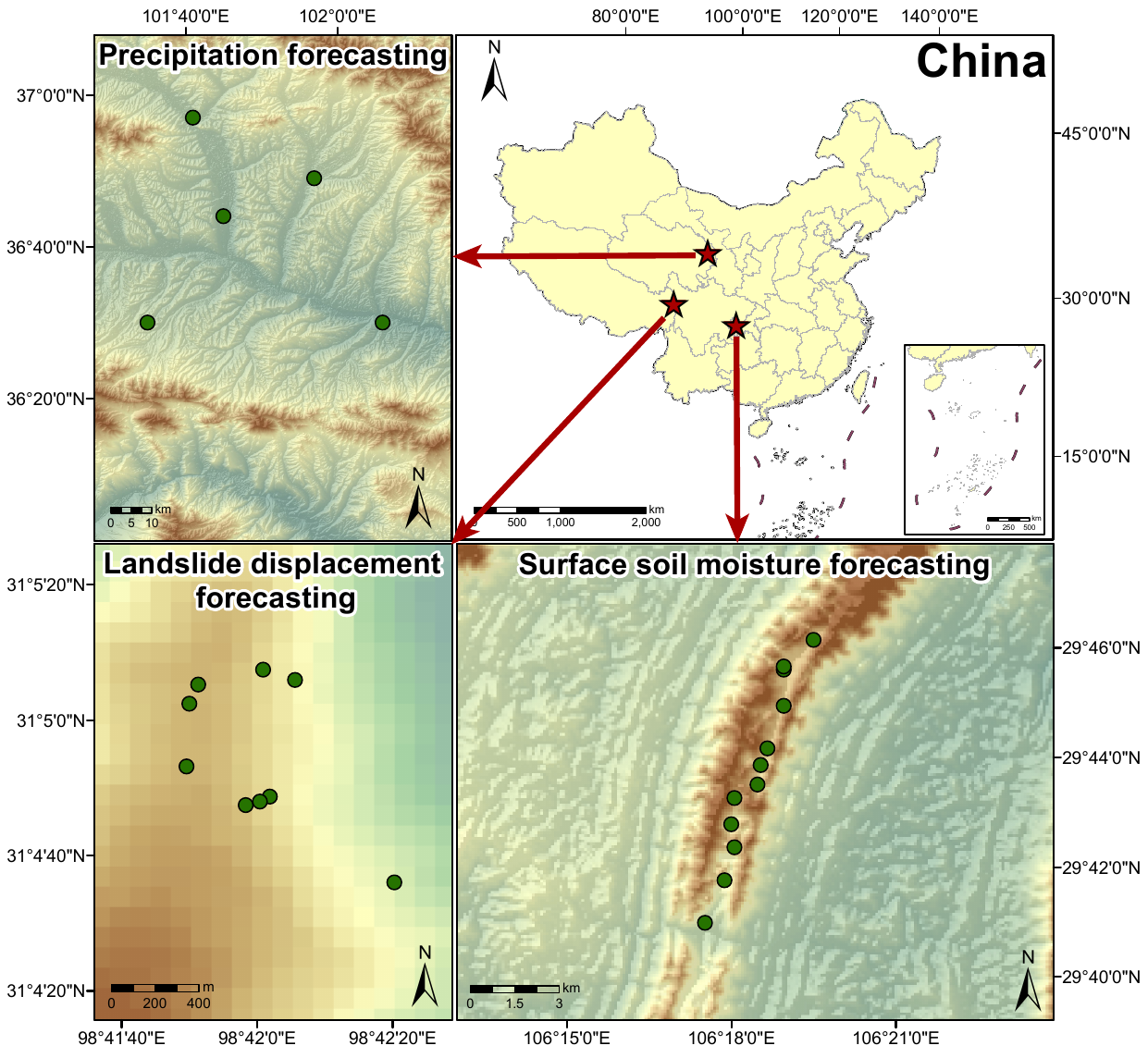}
	\caption{The location of the study areas.} 
	\label{Figure1}
\end{figure}

In this study, we collected observation data from three typical application scenarios to validate the effectiveness of the proposed approach. These datasets comprise: monthly precipitation time-series data from selected monitoring stations around Xining along the Qinghai-Tibet Railway in the northeastern Tibetan Plateau (1997–2017) \citep{RN20}, daily horizontal displacement time-series data from multiple monitoring stations at the Baige landslide (May 26, 2019–July 1, 2022) \citep{RN21,RN22}, and hourly surface soil moisture observation time-series data from multiple monitoring stations in the Qingmuguan watershed of the southwestern mountainous region (April 1, 2020–June 30, 2020) \citep{RN23}. These three datasets encompass different temporal scales, varying inter- station distances, and diverse time-series characteristics, effectively demonstrating the generalizability of the proposed forecasting approach. The geographical locations of the study areas for these three datasets are illustrated in Fig.\ref{Figure1}.

For the collected datasets, we implemented missing value imputation using KNN and linear interpolation methods and temporally aligned the multi-stations time-series. For precipitation time-series data, we selected five meteorological stations (Ping'an, Datong, Huzhu, Xining, and Huangzhong) for predictive analysis. For landslide displacement, we utilized horizontal displacement data from nine monitoring stations at the Baige landslide for forecasting. For soil moisture data, we employed soil moisture measurements at 0–5 cm depth from eight monitoring stations for forecasting.

\subsection{Proposed approach for forecasting of spatiotemporally correlated small Earth data}
\subsubsection{Workflow of the proposed approach}

\begin{figure}[!ht]
	\centering
	\includegraphics[width=13cm]{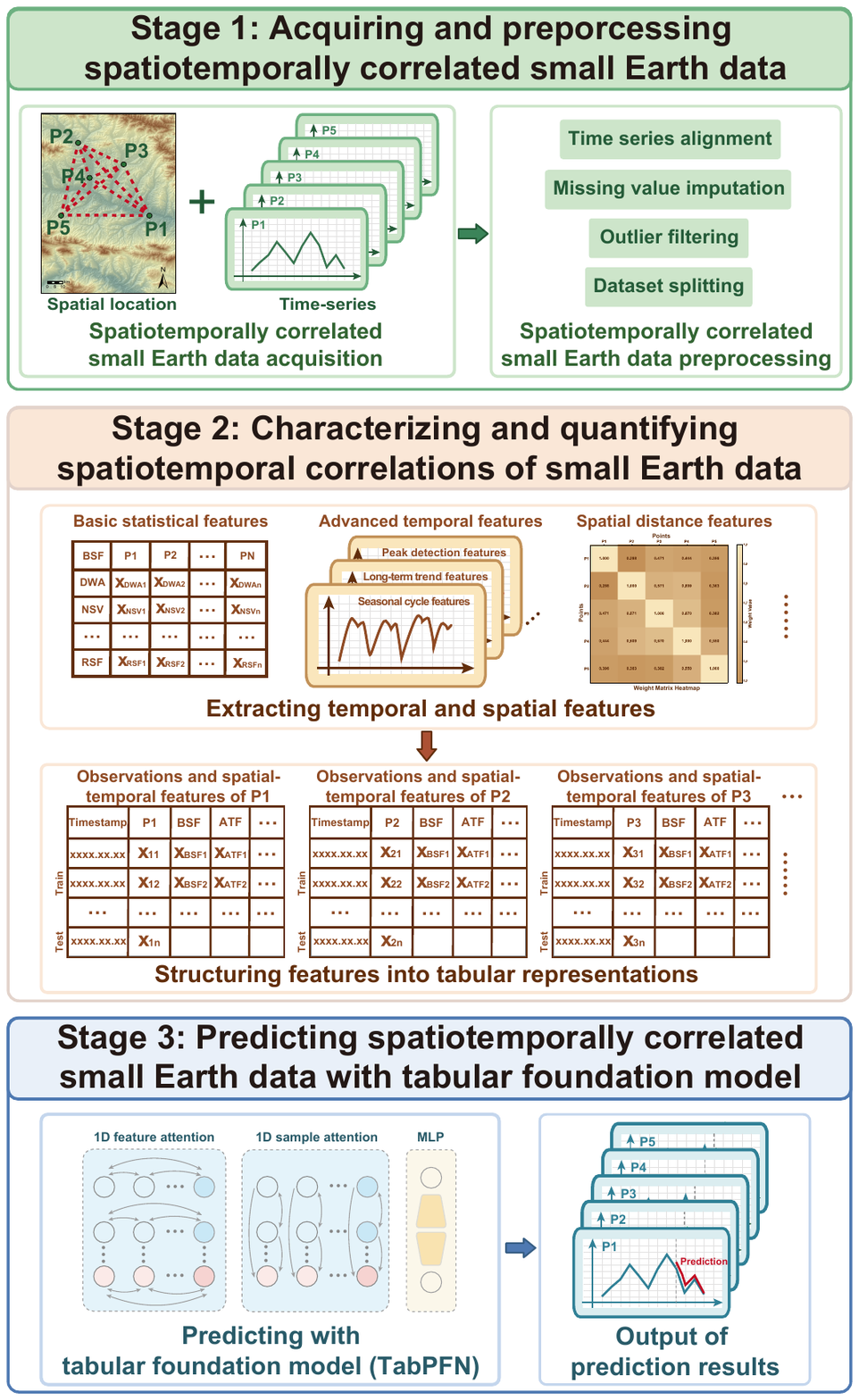}
	\caption{The workflow of the proposed forecasting approach.} 
	\label{Figure2}
\end{figure}

The proposed forecasting approach consists of three stages, (1) spatiotemporally correlated small Earth data are acquired from multiple monitoring stations by collecting their spatial locations and time-series, followed by necessary data preprocessing; (2) the spatiotemporal correlations are then characterized and quantified by extracting temporal and spatial features and structuring them into tabular representations; (3) spatiotemporally correlated small Earth data forecasting is implemented using a tabular foundation model (TabPFN-TS), and the forecasting results are generated and stored for different monitoring locations. The workflow of the proposed forecasting approach is illustrated in Fig.\ref{Figure2}.

\subsubsection{Key idea: characterization and quantification of spatiotemporal correlations}

The essential idea behind the proposed approach of forecasting is to characterize and quantify the spatiotemporal correlations of the small Earth data. The characterization and quantification of spatiotemporal correlations are mainly realized through the extraction of spatiotemporal features, and then the spatiotemporal features are structured into tabular representations to embedding in tabular foundation model to predict.

The spatiotemporal correlations of small Earth data are explicitly characterized and quantified by constructing a comprehensive feature set encompassing temporal and spatial dimensions. Temporal features captured local fluctuations, long-term trends, cyclical variations, and regime shifts, thereby enabling the model to recognize both short-term dynamics and persistent structural changes. Spatial features, including distance-weighted averages, nearest-station effects, and spatial gradients, encoded neighborhood influences and spatial heterogeneity inherent in Earth systems. This feature-based representation is necessary because geoscience data inherently exhibit spatial dependencies across regions and temporal dependencies across scales, which are often neglected by forecasting strategies \citep{RN4,RN24}. By transforming spatiotemporal correlations into explicit numerical representations, the proposed approach enables the model to effectively capture spatiotemporal correlations even under limited data conditions. Furthermore, incorporating these representations into the tabular foundation model (TabPFN-TS) provides additional prior knowledge that reduces overfitting, improves data efficiency, and enhances generalization across monitoring stations \citep{RN25,RN26}.

Basic statistical features provide intuitive information regarding data patterns through five categories: (1) lag features capture autocorrelation properties to characterize temporal dependencies \citep{RN27}; (2) rolling statistics capture local variations and dynamic characteristics to enhance model responsiveness to trend changes; (3) differencing features eliminate trends and stabilize non-stationary time-series data; (4) variability features include Coefficient of Variation (CV) \citep{RN28} and Interquartile Range (IQR) to quantify data stability and dispersion characteristics; (5) cumulative features include Cumulative Sum and Cumulative Ratio to represent aggregate impact and proportional comparisons over temporal windows.

\begin{equation}
	\mathrm{CV}_t^{(w)}=\frac{\sigma_t^{(w)}}{\mu_t^{(w)}+\epsilon}
\end{equation}
where $\sigma_t^{(w)}$ represents the standard deviation on window $\text {w}$, $\mu_t^{(w)}$ represents the mean on window $\text {w}$, and $\epsilon$ is a stabilizing value to avoid division by zero.

\begin{equation}
	\mathrm{IQR}_t^{(w)}=Q 3_t^{(w)}-Q 1_t^{(w)}
\end{equation}
where $\text {Q1}$ and $\text {Q3}$ are the 25th percentile and 75th percentile within the window.

Advanced temporal features identify seasonal, trend, cyclical, and dynamic patterns through five categories: (1) seasonal cycle features use sine and cosine functions to generate continuous cyclical variables that preserve temporal periodicity and enhance recognition of recurring patterns; (2) long-term trend features employ linear or quadratic polynomial fitting to represent sustained directional changes; (3) cyclic group statistics compute mean and standard deviation by temporal units to capture relative anomalies and abnormal fluctuations; (4) peak detection features identify local maxima exceeding the 70th percentile and determine proximity to significant peaks; (5) dynamic window-based features include mean reversion quantification, trend direction indicators, and relative position metrics that normalize current values between recent boundaries to characterize observation positions within local ranges.

Regime change features are primarily employed to capture state transitions and phase changes within time-series data. Two categories of regime change features are implemented. (1) Moving window variance ratio compares short-term volatility (variance) to long-term volatility (variance), thereby detecting recent sudden increases or decreases in volatility patterns, as shown in the equation X. (2) Stage-based statistics divide the time-series into three equal temporal segments: initial, intermediate, and final stages. Average values are calculated for each stage, along with inter-stage change rates, enabling the capture of structural changes across different time periods.

\begin{equation}
	\operatorname{VarianceRatio}_t^{(s, l)}=\frac{\operatorname{Var}\left(x_{t-s+1: t}\right)}{\operatorname{Var}\left(x_{t-l+1: t}\right)+\epsilon}
\end{equation}
where, $x_{t}$ represents the original time-series, s represents the short-term window size, l represents the long-term window size, and $\epsilon$ is the minimum value to prevent division by zero.

Spatial distance features introduce spatial structural information and model spatial dependencies through three categories: (1) distance-weighted averages utilize Gaussian kernel functions with proximity-based weighting where closer stations receive higher weights with exponential decay; (2) nearest station values select top-ranked neighboring stations based on minimum distance and maximum weight criteria, providing both raw and weighted neighboring station values; (3) spatial gradient features calculate rolling average differences between target and neighboring stations within time windows to quantify local value differences and measure spatial variation direction and intensity.

\begin{equation}
	w(d)=\exp \left(-\frac{d}{\sigma}\right)
\end{equation}
where, d represents distance, and $\sigma$ represents the adjustment parameter.

Cross-station dynamic features extract dynamic relationship characteristics between spatially distributed monitoring stations, capturing dynamic correlations, regional consistency, and spatial propagation patterns. Different cross-station dynamic features were selected for different temporal scales. For hourly data, regional synchronicity features were employed, capturing short-term fluctuations and filtering noise in high-frequency data through real-time computation of global states across all stations. This approach demonstrates higher efficiency compared to cross-station statistical features and is better suited for real-time forecasting requirements. For daily and monthly data, cross-station statistical features were utilized, capturing long-term trends and anomalous events through rolling statistics across multiple temporal windows, making them more appropriate for stable forecasting of medium- to low-frequency data.

After extracting the above features and structuring them into tabular representations, intelligent feature selection is applied to embed the features into the tabular foundation model. A hierarchical progressive embedding strategy was adopted for spatiotemporal features embedding, enabling deep fusion between self-defined spatiotemporal features and the intrinsic feature capabilities of TabPFN-TS. First, a spatiotemporally coupled feature framework was constructed by combining universal features with spatial distance features. Then, an intelligent feature selection algorithm was applied to optimize and filter features based on correlation analysis, thereby mitigating redundancy and reducing the risk of overfitting. Finally, a dual adaptive feature mechanism was employed to achieve the fusion of self-defined spatiotemporal features with TabPFN-TS's built-in features, establishing a comprehensive feature system spanning basic statistics, advanced temporal patterns, and spatial distance metrics. This dual feature embedding methodology not only maximizes TabPFN-TS's automatic feature optimization capabilities but also substantially enhances the model's capacity to capture complex temporal patterns through prior-specific innovative feature. The approach achieves enhanced modularity, scalability, and optimized predictive performance, providing a theoretically innovative and practically valuable solution for spatiotemporal forecasting applications.

\subsection{Baselines for comparative evaluation}

\subsubsection{Graph deep learning model (T-GCN) }

This study employs T-GCN \citep{RN11} to perform forecasting on spatiotemporally correlated small Earth data, serving as baselines for comparison to demonstrate the effectiveness of the proposed approach. This model employs the traditional T-GCN architecture optimized for time-series forecasting, utilizing Gaussian transformation to process distance matrices for graph structure construction. The model adopts a two-layer T-GCN structure combined with gating mechanisms (reset gate, update gate, candidate gate) to capture spatiotemporal dependencies. Multi-step forecasting is performed through sliding window forecasting, with training conducted using the Adam optimizer and MSE loss function.

\subsubsection{Tabular foundation model (TabPFN)} 

This study also employs TabPFN \citep{RN18} to perform forecasting on spatiotemporally correlated small Earth data, serving as a baseline. TabPFN represents a type of tabular foundation model based on prior-data fitted networks, proposed by Hollmann et al. \citep{RN18}. Building upon TabPFN's success, researchers further developed TabPFN-TS, a model specifically optimized for time-series forecasting tasks \citep{RN29}. This model employs a Transformer architecture with a pretraining-fine-tuning paradigm, utilizing meta-learning approaches for pretraining on extensive synthetic time-series data to learn universal patterns and representations of time-series, thereby achieving strong generalization capabilities. The core advantage of TabPFN-TS lies in its zero-shot forecasting capability, enabling direct application to novel time-series forecasting tasks without requiring task-specific training. Through specialized temporal position encoding and attention mechanisms, the model effectively captures long-term dependencies, trends, and periodic features within time-series data. TabPFN-TS supports both univariate and multivariate time-series forecasting, demonstrating favorable generalization performance when processing small-sample data. Compared to traditional time-series forecasting approaches, it offers advantages including elimination of complex hyperparameter tuning requirements and rapid adaptation to new data. However, limitations may exist when processing extremely long sequences and certain domain-specific time-series. 

\subsection{Evaluation metrics}

This study employs five metrics to evaluate the performance of forecasting: Mean Absolute Error (MAE), Mean Absolute Percentage Error (MAPE), Mean Square Error (MSE), and Root Mean Square Error (RMSE). Due to the presence of zero values in precipitation data that prevent MAPE calculation, MAPE evaluation is excluded for precipitation forecasting, and the Kling-Gupta Efficiency (KGE) comprehensive evaluation metric is incorporated instead.

(1) Mean Square Error (MSE)

MSE quantifies forecasting accuracy by calculating the average of squared forecasting errors. This metric amplifies larger errors, making it more sensitive to significant deviations during optimization processes. Higher deviation values in forecasting correspond to larger MSE values, indicating poorer model performance. The mathematical expression for MSE is as follows:

\begin{equation}
	\text { MSE }=\frac{1}{m} \sum_{i=1}^m\left(y_i-\hat{y}_i\right)^2
\end{equation}

(2) Root Mean Square Error (RMSE)

RMSE represents the square root of MSE, retaining MSE's sensitivity to large errors while restoring results to the same units as the original data, thereby enhancing interpretability. Smaller values indicate superior model performance. RMSE is commonly employed in hydrology, meteorology, environmental science, and other fields requiring high precision. The mathematical expression for RMSE is as follows:

\begin{equation}
	R M S E=\sqrt{\frac{\sum_{i=1}^n\left(\hat{y}_i-y_i\right)^2}{n}}
\end{equation}

(3) Mean Absolute Error (MAE)

MAE measures the average magnitude of errors between predicted and observed values, representing the mean absolute deviation of forecasting from actual observations. The calculation involves taking the absolute value of all forecasting errors and computing their average. Smaller cumulative errors correspond to lower MAE values. The mathematical expression for MAE is as follows:

\begin{equation}
	M A E=\frac{1}{n} \sum_{i=1}^n\left|\hat{y}_i-y_i\right|
\end{equation}

(4) Mean Absolute Percentage Error (MAPE)

MAPE evaluates forecasting accuracy through relative error analysis, calculating the percentage of error between predicted and observed values relative to observed values, then determining the average. This metric expresses the average percentage deviation. However, MAPE encounters division-by-zero or extremely large value problems when observed values are zero or near zero, necessitating cautious application when processing data containing zero values. Lower MAPE values indicate higher forecasting accuracy. The mathematical expression for MAPE is as follows:

\begin{equation}
	M A P E=\frac{100 \%}{n} \sum_{i=1}^n\left|\frac{\hat{y}_i-y_i}{y_i}\right|
\end{equation}

(5) Kling-Gupta Efficiency (KGE)

KGE represents a comprehensive performance evaluation metric designed to overcome the limitations of traditional Nash-Sutcliffe Efficiency (NSE) regarding sensitivity to bias and variability. KGE provides a more comprehensive assessment of model accuracy, systematic bias, and flow fluctuation simulation capability by simultaneously considering three aspects: correlation ($r$), mean ratio ($\beta$), and variability ratio ($\gamma$). Values closer to 1 indicate superior model performance. Due to its robustness and multidimensional performance considerations, KGE is widely applied in hydrology, water resources, and environmental simulation domains. The mathematical expression is as follows:

\begin{equation}
	\mathrm{KGE}=1-\sqrt{(r-1)^2+(\beta-1)^2+(\gamma-1)^2}
\end{equation}
where $y$ represents observed values, $\hat{y}$ represents predicted values, and $n$ denotes the \underline{amount} of samples. 
$r = \frac{\mathrm{Cov}(\hat{y}, y)}{\sigma_{\hat{y}} \cdot \sigma_y}$ represents correlation, 
$\beta = \frac{\mu_{\hat{y}}}{\mu_y}$ denotes the mean ratio, 
and $\gamma = \frac{CV_{\hat{y}}}{CV_y} = \frac{\sigma_{\hat{y}} / \mu_{\hat{y}}}{\sigma_y / \mu_y}$ indicates the variability ratio. 
$\mu_{\hat{y}}$ and $\mu_y$ are the means of predicted and observed values, respectively, while $\sigma_{\hat{y}}$ and $\sigma_y$ represent the standard deviations of predicted and observed values. 
$\mathrm{Cov}(\hat{y}, y)$ denotes the covariance between predicted and observed values, and $CV_{\hat{y}}$ and $CV_y$ represent the coefficients of variation for predicted and observed values, respectively.

\section{Results and analysis}

We propose an approach for forecasting of spatiotemporally correlated small Earth data with a tabular foundation model and demonstrates its effectiveness across diverse spatial and temporal scales through three representative cases. The first case is monthly precipitation forecasting across different monitoring stations with inter-station distances of approximately 20–60 km. The second case is daily landslide displacement forecasting across multiple monitoring stations with inter-station distances of approximately 40–1,200 m. The third case is hourly surface soil moisture forecasting across various monitoring locations with inter-station distances of 700–8,500 m. We compare the proposed approach with graph deep learning model (T-GCN) and tabular foundation model (TabPFN) to demonstrate its effectiveness and generalizability.

\subsection{Application case 1: monthly precipitation forecasting at different monitoring stations}

\begin{figure}[!ht]
	\centering
	\includegraphics[width=16cm]{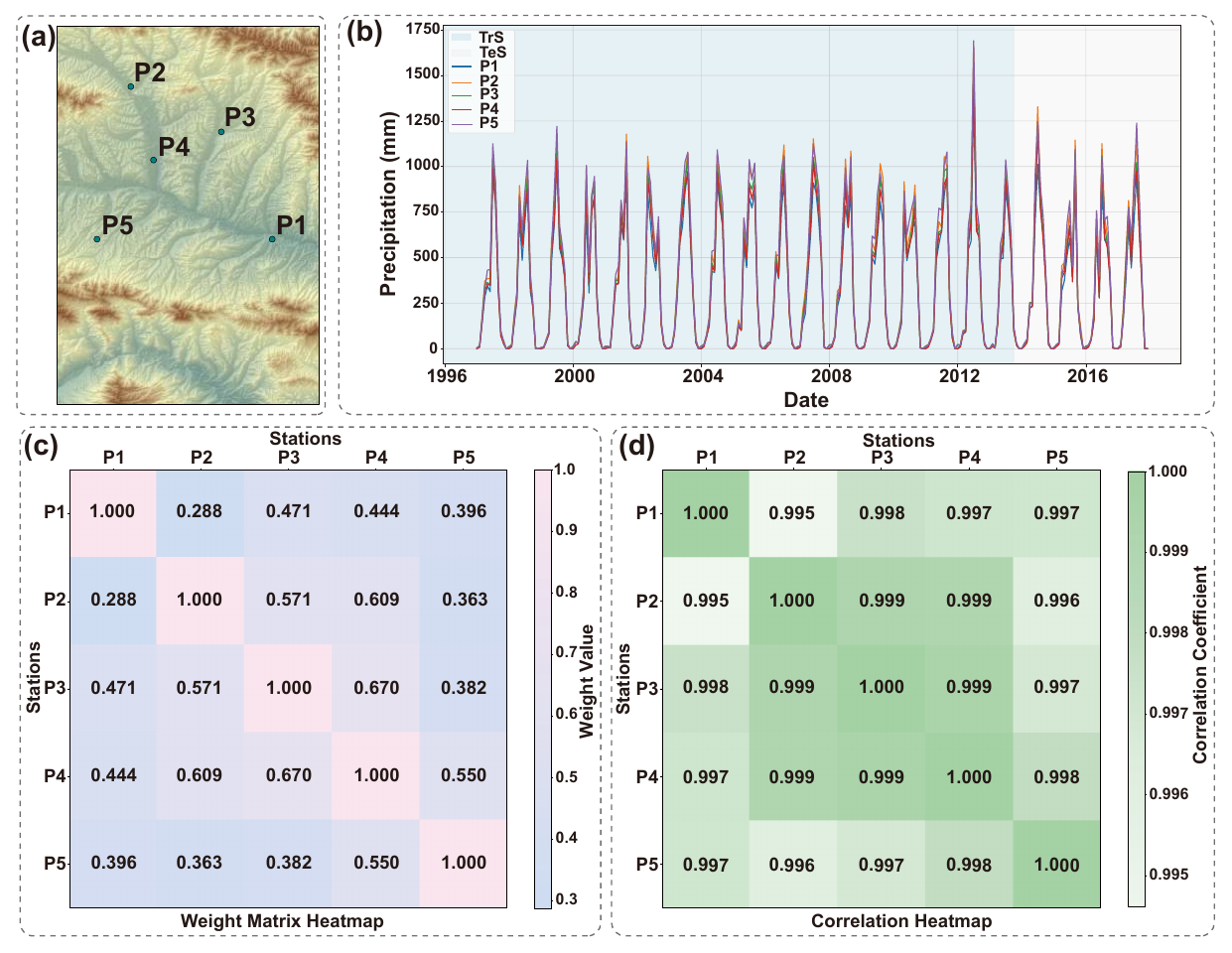}
	\caption{Characteristics and correlations of precipitation monitoring stations. (a) Relative positions of different monitoring stations. (b) Precipitation variations at different monitoring stations. (c) Distance-based weights between different monitoring stations. (d) Correlations of precipitation variations among different monitoring stations.} 
	\label{Figure3}
\end{figure}

The relative positions and inter-station correlations of different precipitation monitoring stations are illustrated through corresponding visualizations and represented via heatmaps, as illustrated in Fig.\ref{Figure3}. The analysis reveals that while the precipitation monitoring stations are geographically dispersed across considerable distances, the precipitation data from different stations exhibit strong correlations that diminish with increasing inter- station distance \citep{RN30}.

\begin{figure}[!ht]
	\centering
	\includegraphics[width=16cm]{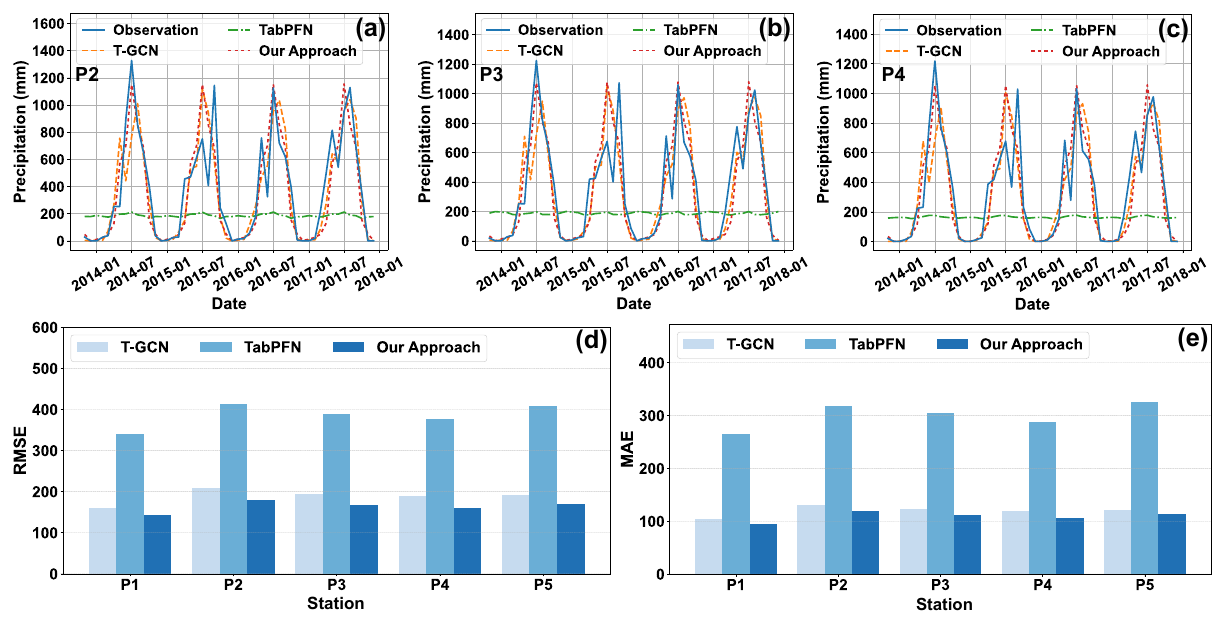}
	\caption{Forecasting performance, (a)–(c) are detailed comparison of precipitation forecasting results between T-GCN based approach, TabPFN based approach and our approach across P2, P3, P4 stations. (d)–(e) are comparison of precipitation forecasting metrics between T-GCN based approach, TabPFN based approach and our approach across different stations.} 
	\label{Figure4}
\end{figure}

We first demonstrated the effectiveness of the proposed approach by applying it in monthly precipitation forecasting at different meteorological stations. The forecasting results for stations P2, P3, and P5 are compared with traditional T-GCN based and TabPFN based approach as illustrated in Fig.\ref{Figure4}(a)–(c). (The overall and detailed precipitation forecasts across all meteorological stations are presented in Supplementary Figure 1–Figure 3.) The data reveal distinct periodicity in precipitation patterns, with both traditional T-GCN based approach and the proposed forecasting approach achieving satisfactory fitting performance. The forecasting effect of the T-GCN based approach is worse than that of the proposed forecasting approach. When employing other monitoring stations as exogenous variables and directly applying the TabPFN-TS model for spatiotemporal forecasting (designated as TabPFN in the figure), the model demonstrates extremely poor predictive performance, with forecasting results exhibiting substantial misalignment. Conversely, the proposed forecasting approach yield satisfactory results, successfully capturing the cyclical fluctuations inherent in the data and achieving robust alignment with observed measurements.

The RMSE and MAE comparison for different monitoring stations is presented in Fig.\ref{Figure4}(d)–(e), (For more detailed comparison of metrics, including MSE and KEG, see Supplementary Figure 3.), with corresponding metric results summarized in Supplementary Table 1. The metric outcomes demonstrate that the proposed forecasting approach achieves optimal predictive performance across all stations. The TabPFN based approach for spatiotemporal data exhibits the poorest predictive performance, with KEG values reaching negative magnitudes, rendering the forecasting results practically unusable. This is because the data from other stations are directly used as exogenous variables. When the numerical differences between the measurement data at different stations are large, the monitoring data from other stations cannot assist in the forecasting, but will interfere with the forecasting effect.

In the precipitation forecasting scenario, although the monitoring sites are spatially distant from each other, the monitoring data exhibit strong correlations across different locations. The characterization and quantification of spatiotemporal features at different sites can effectively contribute to forecasting at all locations, facilitating improved forecasting performance \citep{RN31}. The proposed forecasting method achieves optimal forecasting results at all monitoring sites.

\subsection{Application case 2: daily landslide displacement forecasting at different monitoring stations on the same slope}

\begin{figure}[!ht]
	\centering
	\includegraphics[width=16cm]{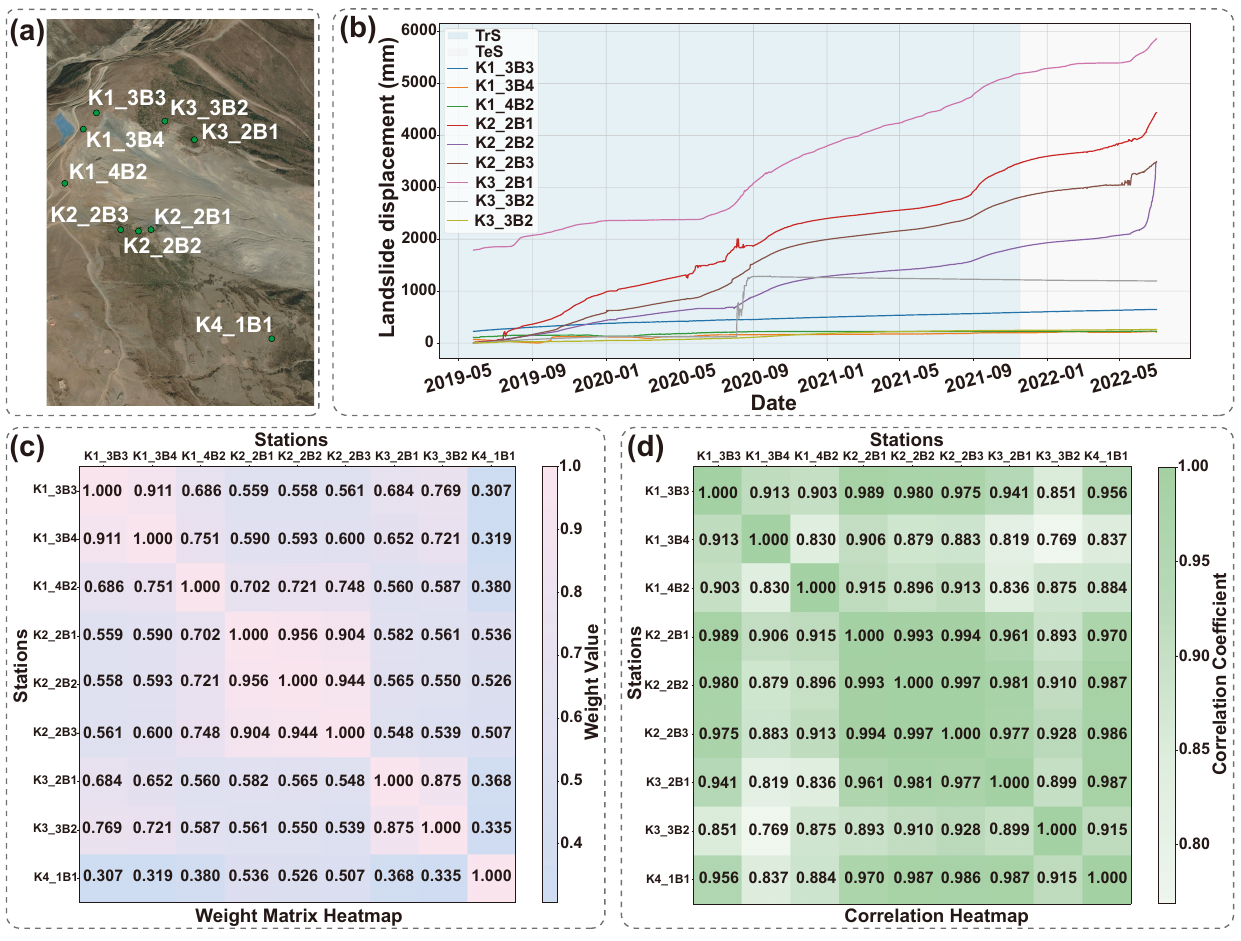}
	\caption{Characteristics and correlations of landslide displacement monitoring stations. (a) Relative positions of different monitoring stations. (b) Precipitation variations at different monitoring stations. (c) Distance-based weights between different monitoring stations. (d) Correlations of landslide displacement variations among different monitoring stations.} 
	\label{Figure5}
\end{figure}

The relative positions and inter-station correlations of different landslide displacement monitoring stations are illustrated through corresponding visualizations and represented via heatmaps, as illustrated in Fig.\ref{Figure5}. The analysis reveals pronounced spatial heterogeneity in landslide displacement patterns, with distinct correlation differentiation among monitoring stations and the presence of multiple deformation pattern clusters. The monitoring stations are strategically distributed across various locations within the Baige landslide: K1-series stations are positioned at the landslide's trailing edge, K2 and K3-series stations are located in the unstable zones of the southern and northern lateral margins respectively, while K4-series stations are situated toward the relatively frontal area of the landslide but maintain considerable distance from the main landslide body.

\begin{figure}[!ht]
	\centering
	\includegraphics[width=16cm]{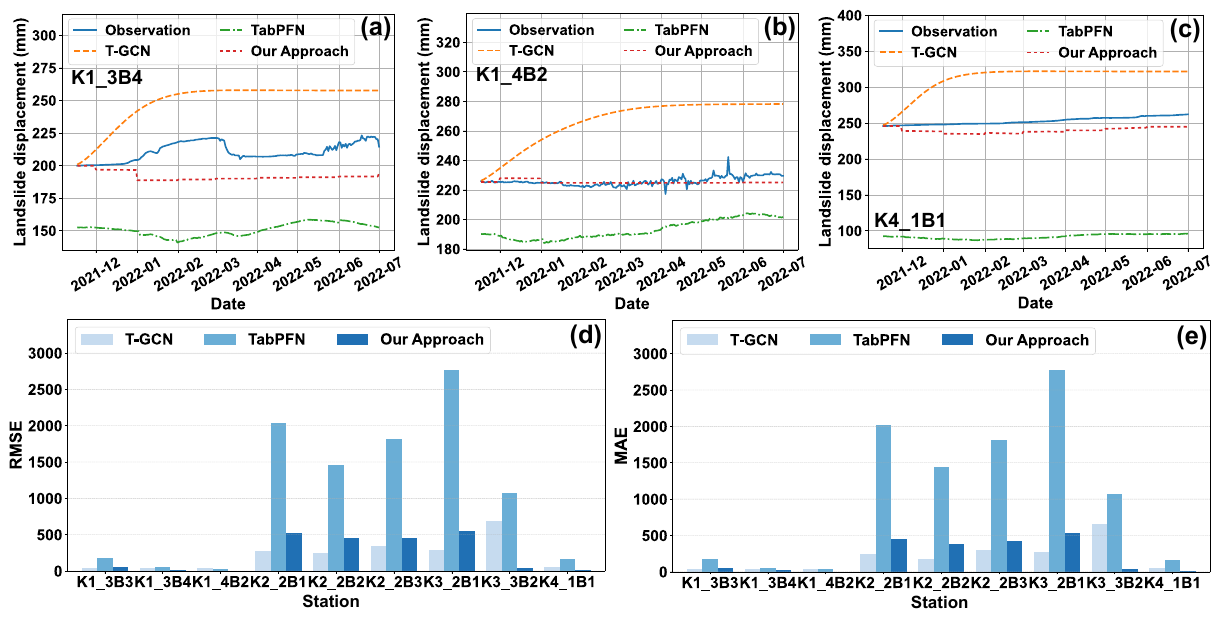}
	\caption{Forecasting performance, (a)–(c) are detailed comparison of precipitation forecasting results between T-GCN based approach, TabPFN based approach and our approach across K1$\_$3B4, K1$\_$4B2, and K4$\_$1B1 stations. (d)–(e) are comparison of precipitation forecasting metrics between T-GCN based approach, TabPFN based approach and our approach across different stations.} 
	\label{Figure6}
\end{figure}

We then demonstrated the effectiveness of the proposed approach by applying it in daily Landslide displacement forecasting at different monitoring stations on the same slope. The forecasting results for stations K1$\_$3B4, K1$\_$4B2, and K4$\_$1B1 are compared with the traditional T-GCN-based approach, TabPFN based approach as illustrated in Fig.\ref{Figure6}(a)–(c). (The overall and detailed precipitation forecasts across all meteorological stations are presented in Supplementary Figures 4–5.) The figures reveal that landslide displacement exhibits no distinct periodic fluctuations but demonstrates a consistent upward trajectory. Notably, displacement trends vary considerably across monitoring stations, posing substantial challenges for predictive displacement. Despite these complexities, the forecasting results demonstrate that while certain stations fail to capture displacement trends accurately, the majority successfully approximate future landslide displacement patterns. The forecasting results demonstrate that the TabPFN-TS forecasting approach for spatiotemporal data continues to yield poor performance, with results completely deviating from the original results and consistently underestimating the predicted values. While the proposed forecasting approach produce relatively satisfactory results. The proposed forecasting approach consistently maintains stable and reliable forecasting performance across all monitoring stations. The proposed forecasting approach consistently produces displacement curves that align closely with observed data, avoiding excessive forecasting deviations (e.g., forecasting of T-GCN based approach for monitoring stations K1$\_$4B2 and  K3$\_$3B2).

The RMSE and MAE comparison for different monitoring stations is presented in Fig.\ref{Figure6}(d)–(e) (for more detailed comparison of metrics, including MSE and MAPE, see Supplementary Figure 6), with corresponding metric results summarized in Supplementary Table 2. The metric outcomes demonstrate that the proposed forecasting approach achieves superior predictive performance across all monitoring stations, with optimal metric values obtained at stations K1$\_$3B4, K1$\_$4B2, K3$\_$3B2, and K4$\_$1B1. Although optimal forecasting accuracy is not achieved at all locations, satisfactory performance is consistently maintained across the remaining monitoring stations. The T-GCN-based approach also demonstrates effective performance in daily landslide displacement forecasting, achieving optimal results at stations K1$\_$3B3, K2-series stations, and K3$\_$2B1. However, poorer forecasting performance is observed at station K1$\_$4B2, which may be attributed to the distinct displacement curve characteristics at this location, exhibiting higher volatility compared to other monitoring stations, while the T-GCN-based approach has limited predictive capability for monitoring sites with high data volatility. These findings further validate the superior generalization capability of the proposed forecasting approach. In contrast, the TabPFN-based approach still exhibits the poorest performance due to significant variations in monitoring values across different sites.

In the landslide displacement forecasting scenario, the displacement patterns at different monitoring stations display unique characteristics, with landslide displacement correlations showing clustering tendencies~\cite{RN32}. K2$\_$2B1, K2$\_$2B2, K2$\_$2B3, and K3$\_$2B1 demonstrate strong inter-correlations, resulting in similar forecasting outcomes. However, temporal similarity in landslide displacement does not constitute the primary determinant of forecasting performance; indeed, station clusters with high temporal correlations (such as the K2 region) exhibit inferior predictive accuracy compared to certain stations with moderate correlations. While the proposed forecasting methodology does not achieve optimal predictive accuracy at certain monitoring locations, it demonstrates the capability to effectively leverage spatiotemporal characteristics for forecasting, thereby achieving satisfactory predictive performance. Notably, at spatial periphery sites such as K4$\_$1B1 (with distance weights $< 0.5$ relative to most other stations), it outperforms alternative approaches~\cite{RN33}, demonstrating the generalizability of the proposed forecasting approach.

\subsection{Application case 3: hourly surface soil moisture forecasting at different monitoring stations}

\begin{figure}[!ht]
	\centering
	\includegraphics[width=16cm]{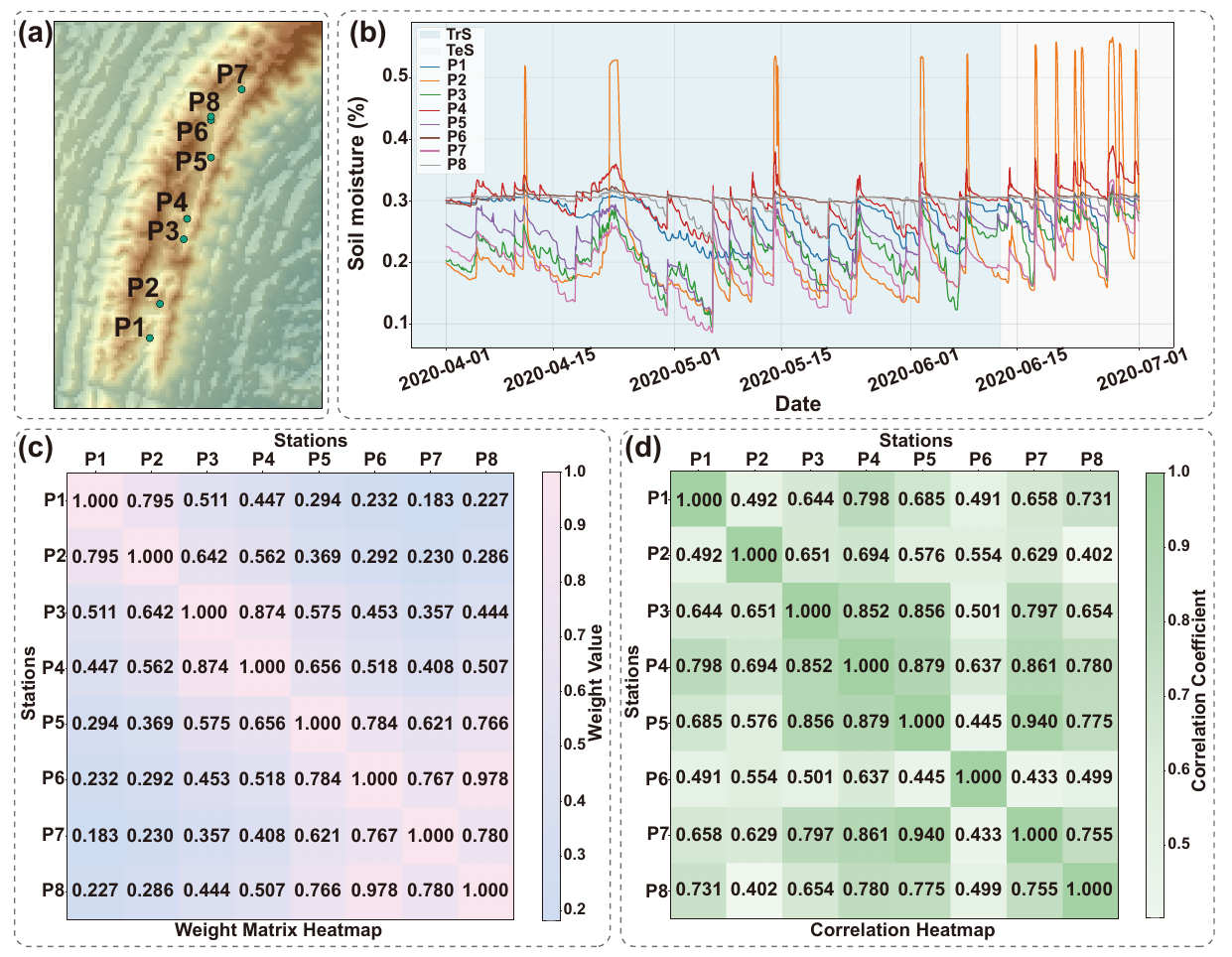}
	\caption{Characteristics and correlations of surface soil moisture monitoring stations. (a) Relative positions of different monitoring stations. (b) Precipitation variations at different monitoring stations. (c) Distance-based weights between different monitoring stations. (d) Correlations of surface soil moisture variations among different monitoring stations.} 
	\label{Figure7}
\end{figure}

The relative positions and inter-station correlations of different surface soil moisture monitoring stations are illustrated through corresponding visualizations and represented via heatmaps, as illustrated in Fig.\ref{Figure7}. The analysis reveals clustering patterns in station positioning, with P1 and P2 in close proximity, P3 and P4 positioned near each other, and P5, P6, P7, and P8 forming a clustered group. Stations with closer spatial proximity acquire more similar temporal and statistical features, consequently yielding comparable forecasting performance \citep{RN34}.

\begin{figure}[!ht]
	\centering
	\includegraphics[width=16cm]{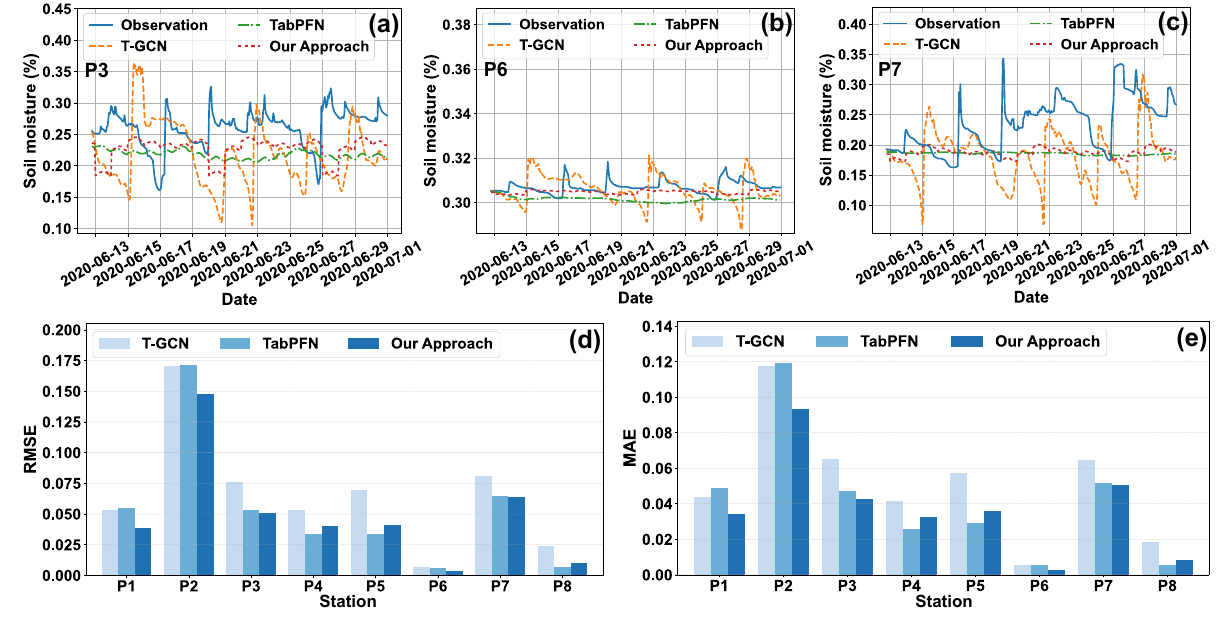}
	\caption{Forecasting performance, (a)–(c) are detailed comparison of precipitation forecasting results between T-GCN based approach, TabPFN based approach and our approach across P3, P6 and P7 stations. (d)–(e) are comparison of precipitation forecasting metrics between T-GCN based approach, TabPFN based approach and our approach across different stations.} 
	\label{Figure8}
\end{figure}

We finally demonstrated the effectiveness of the proposed approach by applying it in Hourly surface soil moisture forecasting at different monitoring stations. The forecasting results for stations P3, P6 and P7 are compared with traditional T-GCN based approach and TabPFN based approach as illustrated in Fig.\ref{Figure8}(a)–(c). (The overall and detailed precipitation forecasts across all meteorological stations are presented in Supplementary Figure 7–Figure 8.) As demonstrated in the figure, soil moisture data exhibits distinct periodic fluctuations. However, substantial variations in fluctuation patterns are observed across different monitoring locations due to site-specific conditions. This variability may introduce interference during spatiotemporal forecasting, consequently compromising predictive accuracy.

The forecasting results reveal that the forecasting of TabPFN based approach generally produce relatively smooth outputs, failing to capture the inherent periodic fluctuations within the data. Conversely, forecasting generated using the proposed forecasting approach successfully capture these fluctuation patterns. Nevertheless, certain forecasting results demonstrate temporal lag relative to observed values, potentially diminishing overall predictive accuracy. The TabPFN based approach achieved relatively satisfactory performance in surface soil moisture forecasting, which is determined by the characteristics of soil moisture monitoring values. Since soil moisture measurements are less than 1 with relatively small differences between monitoring sites, other sites can serve as beneficial exogenous variables for forecasting in certain cases. Conversely, the traditional T-GCN based approach struggles to capture hidden features in monitoring data with pronounced inherent characteristics and high volatility, resulting in significantly inferior predictive performance.

The RMSE and MAE comparison for different monitoring stations is presented in Fig.\ref{Figure8}(d)–(e), (For more detailed comparison of metrics, including MSE and MAPE, see Supplementary Figure 9.), with corresponding metric results summarized in Supplementary Table 3. Metric analysis reveals that the proposed forecasting approach achieving satisfactory predictive accuracy across all stations. Optimal forecasting performance is observed at stations P1, P2, P3, P6 and P7, where the proposed forecasting approach outperforms all baseline approaches across all evaluation metrics. However, inferior forecasting performance is observed at stations P4, P5 and P8 compared to TabPFN based approach. This performance degradation is attributed to temporal lag inherent in the periodic fluctuations predicted by the proposed forecasting approach. Consequently, when predictive effectiveness is evaluated through numerical metrics, the lagged fluctuation patterns produce less favorable results compared to models with smaller fluctuation variations in their predicted outcomes.

In surface soil moisture forecasting scenario, examination of surface soil moisture variations across different monitoring stations (Fig.\ref{Figure7}(b)) and the correlation heatmap (Fig.\ref{Figure7}(d)) indicates generally low similarity in soil moisture changes among various locations. Despite substantial disparities in soil moisture data between different monitoring stations, the implementation of intelligent feature engineering enables effective utilization of alternative features, achieving favorable forecasting performance \citep{RN35}. Notably, traditional T-GCN based approach perform poorly on this type of earth data, exhibiting inferior results at most monitoring stations, while the proposed forecasting approach achieves optimal forecasting performance at the majority of locations, further demonstrating the effectiveness and generalizability of the proposed approach.

\section{Discussion}
\subsection{Simplicity}
The forecasting approach proposed in this study clearly demonstrates simple usability, primarily attributable to its implementation of a tabular foundation model (TabPFN) as the predictive engine. TabPFN represents a deep neural network architecture pre-trained on large-scale heterogeneous datasets, equipped with zero-shot learning capabilities that enable direct and efficient forecasting of multivariate Earth system data without requiring hyperparameter optimization or task-specific model retraining.

Furthermore, the spatiotemporal correlation characterization and quantification in this study operate directly on raw observational data through integrated automated feature generation and selection mechanisms, thereby eliminating the labor-intensive manual feature engineering procedures inherent in conventional approaches \citep{RN36,RN37}. This end-to-end design philosophy endows the approach with exceptional operational feasibility in practical applications: researchers need only input historical observational data sequences to obtain future forecasting of target variables, circumventing complex retraining, parameter tuning, or model selection procedures. This streamlined workflow substantially reduces the implementation threshold while enhancing transferability across diverse earth science forecasting scenarios \citep{RN38,RN39}.

\subsection{Robustness}
In this study, the proposed forecasting approach demonstrates superior predictive performance for spatiotemporally correlated small Earth data forecasting with strong generalizability. For datasets with distinct inherent characteristics, the characterization and quantification of spatiotemporal features enable unique feature strategies tailored to different application scenarios and monitoring locations, thereby further optimizing predictive performance. Furthermore, the utilization of tabular foundation model as the core predictive model enhances the approach's generalizability and operational convenience while reducing developmental costs, providing valuable insights for foundation model applications in geoscience.

The core innovation of this approach lies in its ability to systematically characterize and quantify spatiotemporal features across heterogeneous Earth data systems. The proposed approach captures both local dynamics and cross-spatial dependencies, enabling the model to leverage spatiotemporal correlations \citep{RN40}. The integration of tabular foundation models as the core predictive engine represents a paradigm shift in geoscientific forecasting. By leveraging pre-trained representations and transfer learning capabilities, the approach reduces developmental costs while enhancing operational convenience and model generalizability. This design choice enables the methodology to effectively handle small datasets—a common constraint in Earth system monitoring—while maintaining robust predictive performance across diverse application scenarios.

Validation across multiple geoscientific scenarios reveals that the proposed approach effectively adapts to varying spatiotemporal correlation structures. In scenarios with strong inter-site correlations despite large spatial distances (as observed in precipitation forecasting), the methodology efficiently utilizes cross-spatial information to enhance local predictions. Conversely, in systems with clustered correlation patterns (such as landslide monitoring stations), the approach demonstrates capability to identify and leverage relevant spatial groupings while maintaining performance at spatially isolated monitoring points. Particularly noteworthy is the methodology's performance in scenarios with weak spatiotemporal correlations, such as surface soil moisture systems \citep{RN41}. Traditional graph-based approaches often fail in such contexts due to their reliance on strong connectivity assumptions. The proposed approach overcomes this limitation, demonstrating superior robustness compared to conventional approaches \citep{RN42}.

\subsection{Limitations}
The proposed approach exhibits several limitations. First, different forecasting scenarios contain distinct inherent patterns, and each monitoring location possesses unique independent characteristics. The correlation between different locations is limited, with varying dominant factors influencing forecasting performance. Although the adopted approach achieves satisfactory forecasting results for most stations with some reaching optimal performance, certain stations fail to achieve optimal forecasting accuracy. For instance, in landslide displacement forecasting, since the TabPFN model used in our proposed approach has inherent limitations in predicting linear trends, this somewhat weakens the method's effectiveness in forecasting cumulative landslide displacement. Consequently, while the proposed forecasting approach demonstrates satisfactory overall performance, it exhibits marginally inferior results compared to the T-GCN based approach at certain monitoring locations for landslide displacement forecasting.

Similarly, for precipitation forecasting, more sophisticated and specialized forecasting methods currently exist that may achieve higher accuracy by incorporating comprehensive meteorological variables, atmospheric dynamics, and climate modeling approaches. Our proposed approach may exhibit performance gaps compared to these domain-specific precipitation forecasting models, particularly in terms of the limited range of environmental factors considered. However, precipitation forecasting serves primarily as an application scenario to validate the advantages of our approach of spatiotemporal forecasting rather than competing with specialized meteorological forecasting systems.

Furthermore, the current feature engineering framework focuses exclusively on temporal data characteristics without incorporating environmental information or other enriched exogenous data sources to assist forecasting, which constrains improvements in predictive accuracy. This limitation represents a significant opportunity for methodological enhancement.

\subsection{Future work}
Future research will incorporate more comprehensive exogenous information sources. For surface soil moisture forecasting, topographical and soil type auxiliary data will be introduced as auxiliary information, while landslide displacement forecasting will benefit from rainfall lag variables and geological classification information. These enhancements will improve the accuracy and effectiveness of forecasting results. Additionally, predictive methodologies can be further optimized by introducing spatial clustering features for small Earth data exhibiting pronounced spatial aggregation patterns. The implementation of spatial clustering combined with local modeling approaches will enhance adaptability to complex local structures, enabling more accurate forecasting.

\section{Conclusion}

This study presents a simple and robust approach for forecasting of spatiotemporally correlated small Earth data, with validation conducted across three application scenarios: precipitation forecasting, landslide displacement forecasting, and surface soil moisture forecasting. The validation demonstrates the effectiveness and generalizability of the proposed approach. The approach characterizes and quantifies spatiotemporal patterns by extracting inherent spatial and temporal features across different scenarios and employ tabular foundation model as the core forecasting architecture, thereby establishing an effective forecasting approach for spatiotemporally correlated small Earth data forecasting. The research findings demonstrate that the proposed approach achieves superior predictive performance compared to the graph deep learning model (T-GCN) and tabular foundation model (TabPFN) in the majority of instances, exhibiting strong robustness.

Future research will focus on incorporating more comprehensive multi-source spatiotemporal and environmental information for different application scenarios within geoscience. The integration of spatial clustering and local modeling modules will establish more robust deep learning prediction methodologies with superior performance characteristics. Additionally, the approach holds potential for broader application in other engineering and scientific domains requiring efficient time series forecasting capabilities.

%

\section*{Acknowledgments}
This research was jointly supported by the National Science Foundation of China (Grants No. 42277161), the China Postdoctoral Science Foundation (Grant No. 2024T170859), and the Postdoctoral Fellowship Program of CPSF (Grant No. GZB20230685). During the preparation of this work, the authors used AI-based tools to improve language clarity. After using these tools, the authors reviewed and edited the content as needed and took full responsibility for the content of the publication.

\bibliographystyle{elsarticle-harv} 
\bibliography{reference}


\end{document}